\documentclass[letterpaper]{article} 
\usepackage{aaai25}  
\usepackage{amsmath}
\usepackage{color}
\usepackage{times}  
\usepackage{helvet}  
\usepackage{courier}  
\usepackage[hyphens]{url}  
\usepackage{booktabs}

\usepackage{listings}
\usepackage{graphicx} 
\usepackage{natbib}  
\usepackage{caption} 
\usepackage{algorithm}
\usepackage{algorithmic}

\usepackage{newfloat}
\usepackage{listings}
\DeclareCaptionStyle{ruled}{labelfont=normalfont,labelsep=colon,strut=off} 
\floatstyle{ruled}
\newfloat{listing}{tb}{lst}{}
\floatname{listing}{Listing}
\title{MEPG: Multi-Expert Planning and Generation for Compositionally-Rich Image Generation}

\author{
    Zhao Yuan\textsuperscript{\rm 1}, 
    Liu Lin\textsuperscript{\rm 2}\thanks{Corresponding author} \\
}
\affiliations{
    \textsuperscript{\rm 1}CCMU, China \\
    \textsuperscript{\rm 2}University of Science and Technology of China,China \\
}

\usepackage{bibentry}

\begin{document}

\maketitle

\begin{abstract}
Text-to-image diffusion models have achieved remarkable image quality, but they still struggle with complex, multielement prompts, and limited stylistic diversity. To address these limitations, we propose a Multi-Expert Planning and Generation Framework (MEPG) that synergistically integrates position- and style-aware large language models (LLMs) with spatial-semantic expert modules. The framework comprises two core components: (1) a Position-Style-Aware (PSA) module that utilizes a supervised fine-tuned LLM to decompose input prompts into precise spatial coordinates and style-encoded semantic instructions; and (2) a Multi-Expert Diffusion (MED) module that implements cross-region generation through dynamic expert routing across both local regions and global areas. During the generation process for each local region, specialized models (e.g., realism experts, stylization specialists) are selectively activated for each spatial partition via attention-based gating mechanisms. The architecture supports lightweight integration and replacement of expert models, providing strong extensibility. Additionally, an interactive interface enables real-time spatial layout editing and per-region style selection from a portfolio of experts. Experiments show that MEPG significantly outperforms baseline models with the same backbone in both image quality and style diversity.
\end{abstract}

%
\begin{figure}[ht]
    \centering
    \includegraphics[width=1\linewidth]{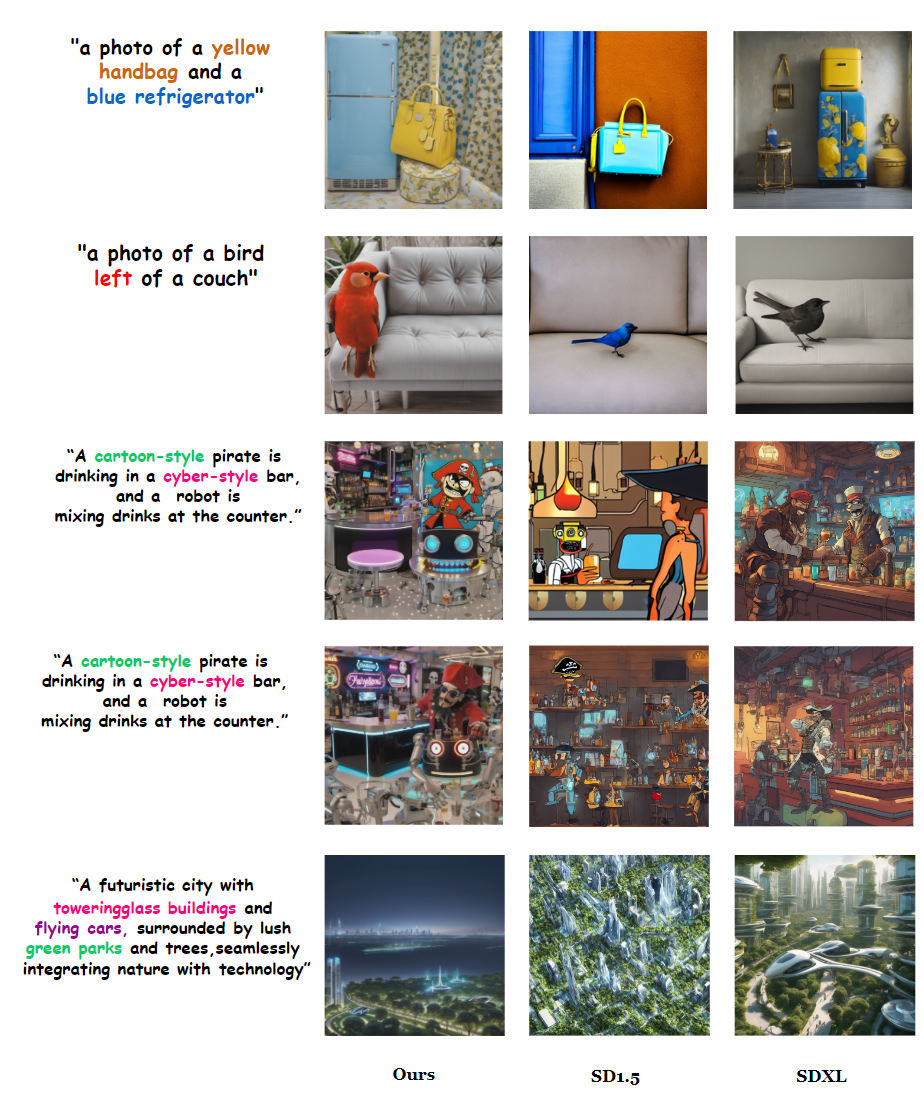}
    \caption{ Compare MEPT with other models. The first two groups of pictures demonstrate the accuracy of generation, while the last three groups of pictures illustrate the diversity of styles. }
    \label{fig:enter-label}
\end{figure}
\section{1 Introduction}\vspace{0.2cm} %
Recent advance in diffusion-based models have demonstrated remarkable progress in generative tasks \cite{Sohl-Dickstein,ho2020denoising,nichol2021improved,dhariwal2021diffusion,saharia2022photoreal}, with notable achievements in text-to-image (T2I) synthesis exemplified by models such as DALL·E 3~\cite{Betker}, Flux~\cite{Batifol}, and Stable Diffusion 3~\cite{Esser}. While the visual fidelity of generated images continues to improve, these models often struggle to follow complex prompts.

To address challenges in multi-object, multi-attribute relational reasoning, current research primarily employs two approaches.
The first leverages image understanding feedback for diffusion optimization, though such methods require high-quality feedback collection and incur additional training overhead~\cite{fan2023reinforcement,black2023training,lee2023aligning,liang2024step,yangdense}. The second incorporates explicit layout conditioning, as demonstrated by prior work utilizing predefined spatial frameworks to enhance local structural coherence~\cite{xie2023boxdiff,yang2023reco,zheng2023layoutdiffusion,li2023gligen}. However, existing layout-conditioning generation methods contains three critical limitations: (1) Existing spatial guidance through layout or attention mechanisms operates at coarse granularity, impeding precise control over object scale and placement; (2) Layout conditioning frequently compromises global aesthetic quality, manifesting as reduced lighting/shadow realism and stylistic diversity; (3) Single-model architectures struggle to balance diverse spatial and semantic token generation requirements across heterogeneous regions, further constraining output quality and diversity.


Additionally, the Mixture of Experts (MoE)~\cite{jacobs1991adaptive} paradigm offers an effective approach for multi-object, multi-attribute generation by scaling model capacity. This method employs dynamic routing mechanisms to selectively activate task-relevant expert sub-networks, enhancing adaptability and output diversity. Nevertheless, conventional MoE implementations~\cite{NEURIPS2023_821655c7,Wan_Video_Wan2_2,mole,balaji2022ediffi} rely solely on textual features or denoising timesteps for expert selection, lacking explicit spatial-semantic reasoning capabilities to enable fine-grained expert specialization.

To reconcile spatial precision with generative diversity under complex textual descriptions while overcoming traditional MoE limitations, we propose MEPT (Multi-Expert Planning Text-to-Image Generation), a novel framework integrating large language model (LLM)-derived spatial coordinates with spatially-aware expert allocation. MEPT establishes a dual-conditioning architecture that combines textual prompts with localized spatial-semantic specifications. Through our designed cross-diffusion mechanism, the framework enables coordinated local-global generation by dynamically selecting optimal experts for each spatial region. Additionally, we implement an editable interface supporting user-adjustable regional layouts and descriptions, significantly enhancing controllability and customization.

Extensive quantitative and qualitative evaluations across multiple benchmarks demonstrate MEPT's superior performance over state-of-the-art methods (e.g., Flux 1.1) in spatial accuracy (+12.3\% FID improvement), diversity (+18.7\% LPIPS gain), and aesthetic quality. Our principal contributions include:

\begin{enumerate}
    \item A novel T2I framework synergizing LLM-based spatial planning with multi-expert collaborative generation;
    \item A cross-diffusion mechanism ensuring global-local generation consistency;
    \item A modular architecture supporting plug-and-play expert integration across diffusion models;
    \item Comprehensive empirical validation of MEPT's effectiveness in complex text-to-image scenarios.
\end{enumerate}

\section{2 Related Work}
\vspace{0.2cm} %
\subsection{2.1 Diffusion-based Model for T2I Generation}
\vspace{0.2cm} %
Diffusion models generate images through a progressive denoising process~\cite{ho2020denoising}, with the training objective of minimizing noise prediction error. Compared to autoregressive models (e.g., LlamaGen~\cite{sun2024autoregressive}), diffusion models (e.g., DALL-E 3~\cite{Betker}, FLUX~\cite{Batifol}, and SD 3~\cite{Esser}) exhibit superior compositional generation capabilities. For instance, on some T2I benchmarks~\cite{ghosh2023geneval,hu2024ella}, diffusion models significantly reduce entity missing and incorrect attribute binding errors. 
The model we propose is based on a diffusion model that utilizes LLM and MOE strategies to improve existing text-to-image approaches.

\subsection{2.2 Mixture of Experts}
With the continuous advancement of artificial intelligence, modern datasets have become increasingly diverse and complex~\cite{lin2014microsoft,changpinyo2021conceptual,schuhmann2022laion5b}, which makes it increasingly difficult for a single model to integrate conflicting or heterogeneous knowledge. One approach to address this is the expert mixture of experts (MOE) architecture~\cite{jacobs1991adaptive,shazeer2016outrageously}.
The MOE model dynamically selects and activates the appropriate experts based on the input data features, thereby achieving an optimal balance between performance and efficiency.
In recent years, MOE has achieved significant success in the field of NLP~\cite{fedus2022switch,deepseekmoe,du2022glam}. Recently, MOE has also made progress in the area of image/video generation. And these methods can be divided into Textual-Feature Routing and Denoising-Timestep Routing.  The former~\cite{Xue2023Raphael} dynamically allocates expert resources based on the semantic features of the input text (such as word embeddings and contextual representations), while the latter~\cite{Wan_Video_Wan2_2,balaji2022ediffi,NEURIPS2023_821655c7} dynamically allocates experts according to the time steps of the denoising process.
Unlike these methods, our approach possesses spatial-semantic reasoning capabilities and enables fine-grained expert specialization.


\subsection{2.3 Layout Conditioning Generation}
\begin{figure*}[htbp] 
\begin{center}
    \includegraphics[width=1\linewidth]{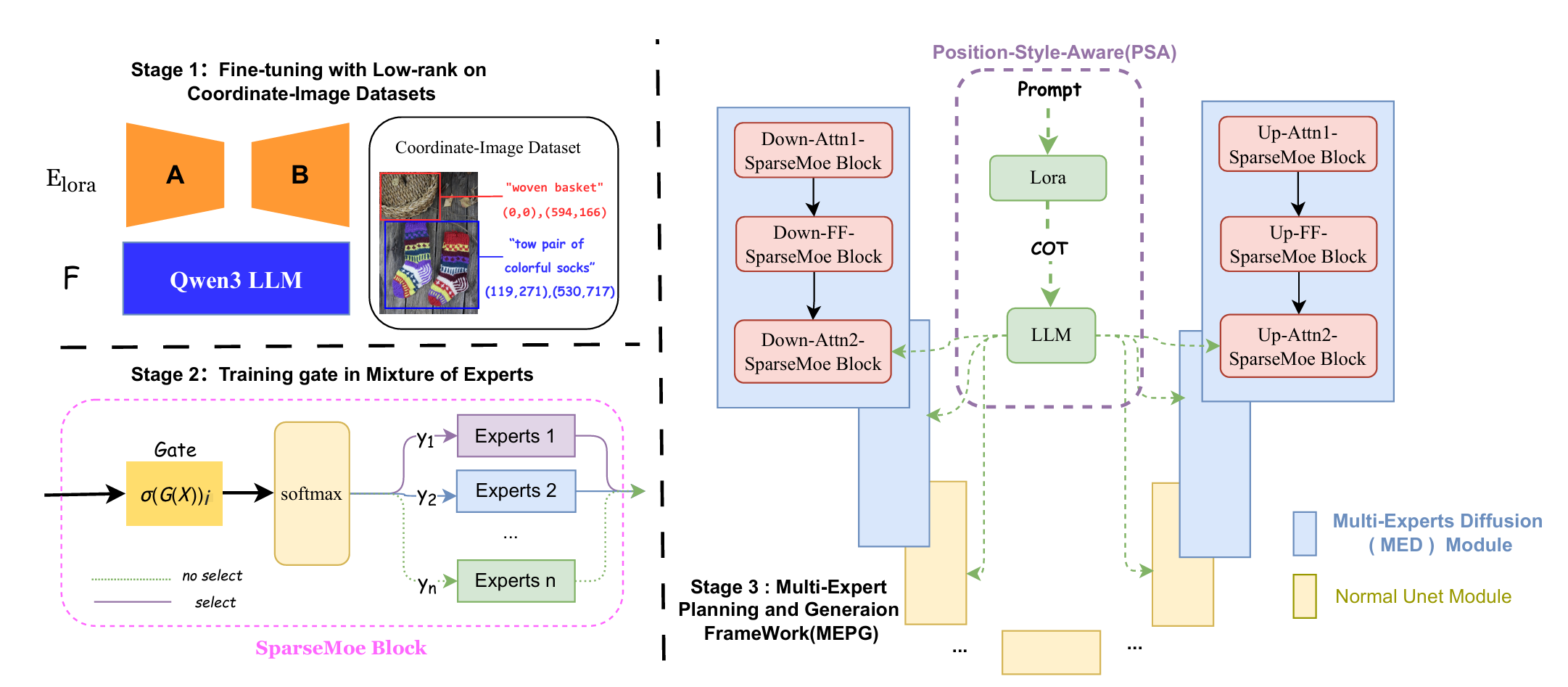}
\end{center}
    \caption{The overall framework of our proposed Multi-Expert Planning and Generation (MEPG), which contains three stages: (1) Fine-tuning with Low-rank on
Coordinate-Image Datasets. (2) Training gate in Mixture of Experts. (3) Multi-Expert
Planning and Generation.}
    \label{fig:2}
\end{figure*}
Recent research in conditional or controllable image generation has demonstrated the effectiveness of explicitly incorporating layout information to guide the synthesis process and improve local structural coherence. 
It can be termed layout conditioning generation, which leverages predefined spatial frameworks (bounding boxes or segmentation masks) to serve as strong constraints during generation.
For example, Xie et al.~\cite{xie2023boxdiff} utilize boxes for compositional control; Yang et al. introduced ReCo~\cite{yang2023reco} for regional control of any object described by open-area text; Zheng et al. presented LayoutDiffusion~\cite{zheng2023layoutdiffusion}, explicitly conditioning diffusion models on layouts; and GLIGen~\cite{li2023gligen}, developed by Li et al., injects spatial grounding into new trainable layers via gated mechanisms. These works enhance the model's ability to generate images where objects adhere faithfully to desired positions, scales, and inter-object relationships.
LayoutGPT\cite{Feng2023layout} improves the generation effect by providing some additional spatial conditions, and RPG \cite{yang2024mastering} uses large language models to process prompts and generates images through cross-attention diffusion.
While existing layout-conditioning methods suffer from imprecise object control, compromised aesthetic realism, and single-model architectural limitations that constrain quality and diversity. Our approach overcomes these critical challenges by integrating a fine-tuned Large Language Model (LLM) and a Mixture-of-Experts strategy, enabling semantically-rich spatial grounding and specialized heterogeneous generation.








\section{3 Method}\vspace{0.2cm} %
\subsection{
3.1 Complex Text Prompt Processing
 }\vspace{0.2cm} %
 
 During image generation, humans naturally consider the size and spatial layout of objects. In contrast, most current models lack an explicit reasoning process when handling prompts. Inspired by Chain-of-Thought (CoT)~\cite{wei2023chain} reasoning in language models, we designed template instructions for complex text prompts containing multiple entities and relationships. These templates guide large language models to analyze the elements, match corresponding attributes, and construct guiding questions to further build spatial layouts. 
 
To simplify the problem, we abstract the spatial position as a rectangular box represented by the two coordinates $(x1,y1)$ and $(x2,y2)$, and assume that the values of x and y are within range$ [0, 1000)$, enabling precise control over the position of visual elements. Based on the textual prompts, we further optimize details, supplement missing descriptions, and combine both into an accurate prompt with precise spatial location information and detailed descriptions. This process is shown in Figure 3.

However, existing large language models are limited by their emergent textual abilities and lack sufficient spatial reasoning. When faced with complex prompts, they often fail to extract the correct elements and may inaccurately describe the size and position of objects (see Figure 3 for comparison). To address this, we trained an auxiliary model using Low-Rank Adaptation (LoRA)~\cite{hu2021lora} to assist the language model in more accurate processing, forming our PSA module.

The prompt processing example in Figure 3 demonstrates how our LoRA model collaborates with the LLM. First, the LoRA model analyzes and processes the original prompt, identifies the constituent elements and their spatial relationships, and passes the processed result as a “Better Prompt” to Step One. Then, leveraging its trained ability to accurately capture object size and spatial layout, the LoRA model converts the prompt into spatial coordinate pairs, which are passed to Step Two for correction of spatial position and size. Finally, the optimized and enhanced information is passed to Step Three, resulting in the final information used for multi-expert model generation.
This information is then processed and converted into prompt-mask pairs for subsequent generation.
\begin{figure*}[htbp] 
\begin{center}
    \includegraphics[width=1\linewidth]{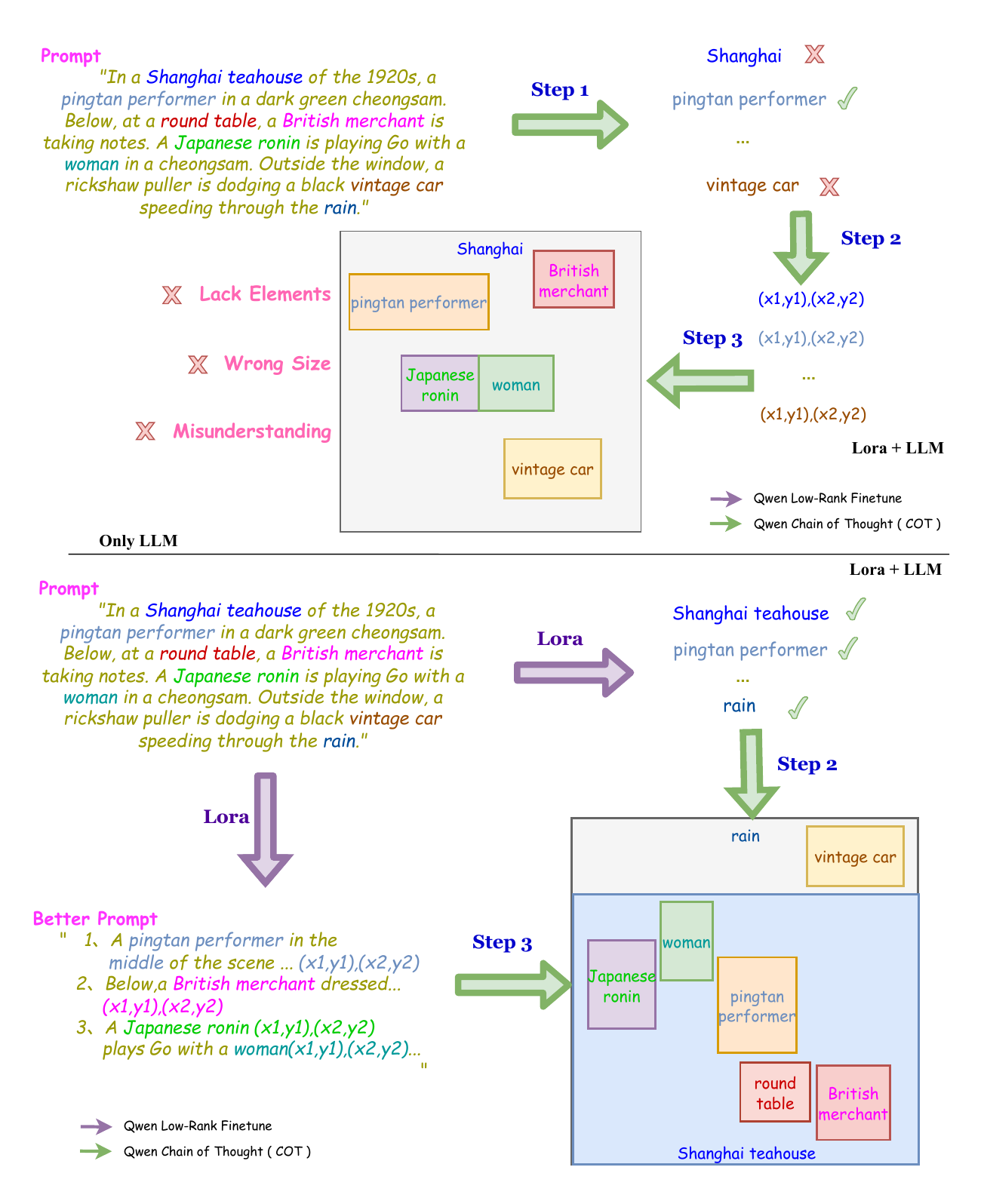}
\end{center}
    \caption{A comparison between using only the LLM chain of thought and using Low-rank Model-assisted LLM chain of thought: (1) It's just that there are problems such as missing elements when using large models, misunderstandings in description, and inaccurate layout of positions. (2) Our Low-rank Model-assisted LLM chain of thought.}
    \label{fig:3}
\end{figure*}


\subsection{\textbf{3.2 Mixture of Experts Diffusion ( MED )  Module}}\vspace{0.2cm} %
Unlike previous MOE applications in image generation methods that set experts for denoising timesteps or assign an expert for each type of entity object, we set a certain number of experts and then train a gating mechanism to activate a specific number of experts during the inference phase.
In the Mixture of Experts Diffusion module, we designed a SparseMoeBlock to integrate various expert models. As shown in Figure 2, this block consists of two parts: a gating function, a softmax function, and the expert modules. The gating function assigns weights to the multiple expert models based on the input data, and these weights are used to adjust the activation of different expert models.
The mathematical formula can be represented as:
$$w_i = \sigma(G(X))_i$$
where $w_i$ represents the weight assigned to the $i$-th expert model;
   $\sigma$ is the sigmoid function;
 $G(X)$ is the output of the gating function applied to the input $X$ (usually a scalar);
 $\sigma(G(X))_i$ is the $i$-th element of the output of the sigmoid function applied to $G(X)$, representing the weight assigned to the $i$-th expert model.


Next, the final output is the weighted sum of each expert, as shown above:
$$y = \sum_{i=1}^{n} w_i \cdot Expert_i(X)$$
where $y$ and $Expert_i$ is the final output and the $i$-th expert model, respectively.

Through the weight allocation mechanism of the SparseMoeBlock, the gating function dynamically controls the participation of each expert model. We replace the qkv components of both the feed-forward (FF) and attention (Attn) modules in denoising UNet with our SparseMoeBlock, where the experts are stacked from the FF or Attn modules of several base models, enabling model selection and weight distribution.
Since our MED module does not modify the original model parameters, it supports the lightweight addition and removal of expert models.





\subsection{
3.3 Cross Denoising Scheduling
}\vspace{0.2cm} %
\begin{figure}[ht]
    \centering
    \includegraphics[width=1\linewidth]{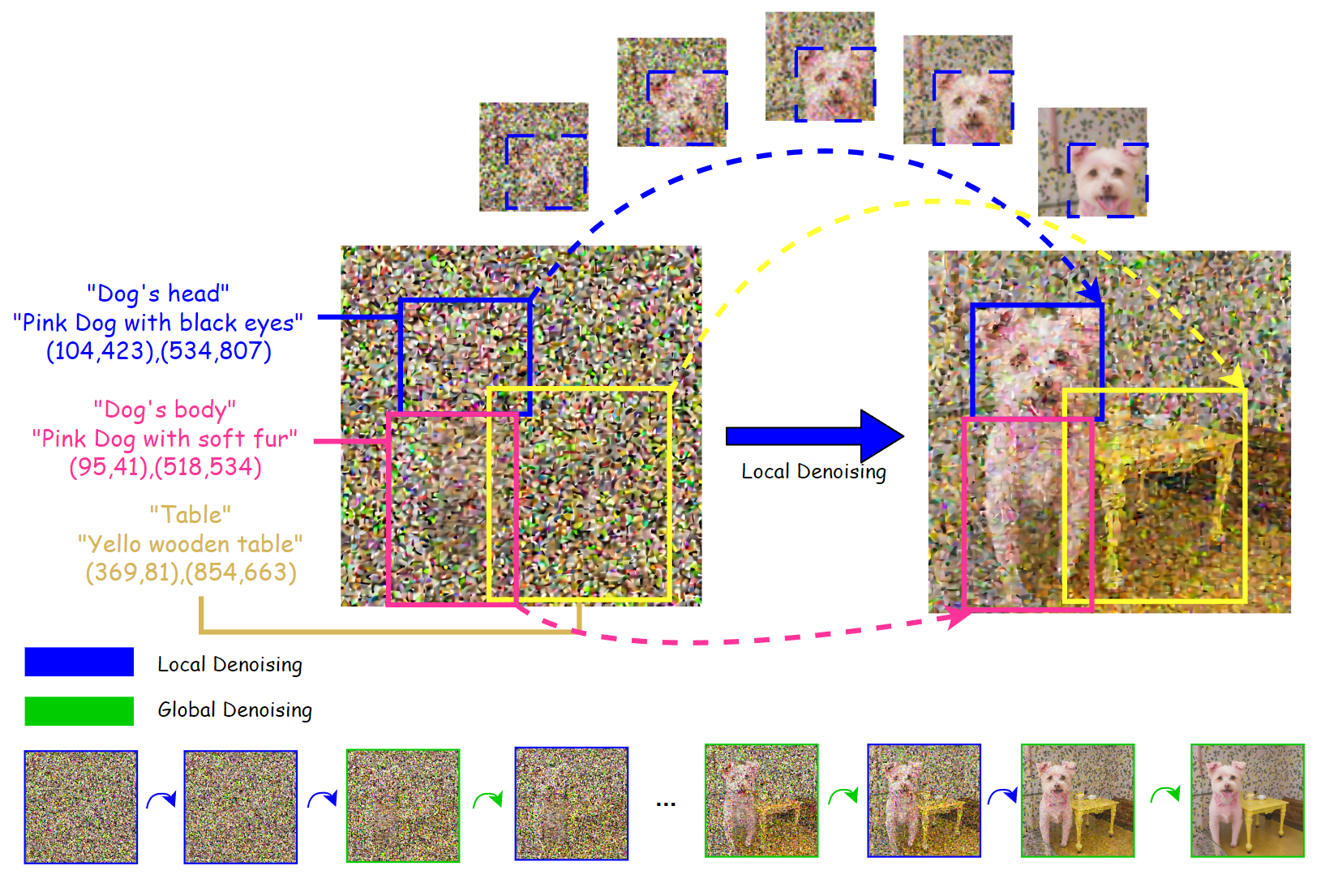}
    \caption{The process of cross-denoising. The following section of denoising demonstrates how we arrange global denoising and local denoising throughout the entire denoising process. The above section demonstrates how to perform Local denoising based on prompts.}
    \label{fig:4}
\end{figure}

To balance global consistency and local detail control in image generation, we propose a cross denoising mechanism that combines global and local denoising strategies,which have been shown in figure 4 . Given a total of $N$ denoising steps, we divide the process into two phases: the first $p_{1} $ proportion of steps are dominated by local denoising, and the remaining $1-p_{1}$ proportion are dominated by global denoising.Specifically,let $t$ denote the current denoising step.The denoising stage at step $t$ is scheduled as follows:

\begin{equation}
\text{Stage}(t) =
\left\{ 
\begin{array}{@{}l@{\quad}l} 
\text{Local denoising}, & t \leq p_1 N \\ 
\text{Global denoising}, & t > p_1 N 
\end{array} 
\right. 
\end{equation}
During the local denoising phase, we partition the latent variable into multiple local windows, each associated with a specific prompt and processed by a corresponding expert sub-model. This enhances the model's ability to generate fine-grained local content. Although global denoising and local denoising have stage divisions, there is still a certain degree of interweaving, global denoising is interleaved to maintain overall structural alignment.

In the global denoising phase, the global expert takes the lead in generation, while local experts provide auxiliary detail refinement. The outputs from all experts are fused using a weighted average at each step:

\begin{equation}
x_t = \sum_{i=1}^M \alpha_i^{(t)} x_t^{(i)}
\end{equation}
where $x _{t} ^{(i)}$  denotes the output from expert $i$ at step $t$, $\alpha _{i} ^{(t)}$ is its corresponding weight, and $M$ is the number of expert models.
This cross denoising scheduling effectively balances global semantic consistency and local detail generation, allowing each region to benefit from the strengths of different expert models and enabling fine-grained control and diverse outputs.

\section{4 Experiments}\vspace{0.2cm} %

\subsection{4.1 Experimental Setup}\vspace{0.2cm} %
\noindent\textbf{Stage 1.}
The based model we choose is Qwen3-8B~\cite{qwen3} and the added LORA is trained using the publicly available Laion-Aesthetics-High-Resolution-GoT (LAHR) dataset~\cite{fang2025got}. This dataset contains 3.77 million high-quality annotated images, with an average prompt length of 110.81 characters per sample. It not only provides richer and optimized annotations for the original prompts but also includes precise spatial coordinate-element pairs (Several examples are shown in Appendix).

\noindent\textbf{Stage 2.}
For gating training of the Multi-Experts-Diffusion (MED), we used the mengcy/LAION-SG dataset~\cite{li2024laion}, which includes 540,000 images in various styles with aesthetic ratings greater than 6. We froze the parameters of the other expert models and trained only the Gate function in the SparseMoeBlock, enabling it to correctly allocate weights to each model.
We selected four models based on the SDXL architecture as the expert group, with some experts better suited for generating realistic styles, and others excelling in generating anime styles. For example, SG161222/RealVisXL is more suitable for photo-realistic image generaion .

\noindent\textbf{Stage 3.} In the inference, we set the total number of denoising steps $N=50$ and the local-dominant proportion $p_{1}=0.7$, meaning that the first 35 steps focus on local denoising and the last 15 steps focus on global denoising. The fusion weights $\alpha _{i} ^{(t)}$ are empirically determined.

\begin{table*}[ht]
    \centering
    \caption{Comparison of Models on Various Metrics on Geneval.}
    \label{tab:comparison}
    \begin{tabular}{l c c c c c c c}
        \toprule
        \textbf{Model} & \textbf{Overall} & \textbf{Single object} & \textbf{Two object} & \textbf{Counting} & \textbf{Colors} & \textbf{Position} & \textbf{Color attribution} \\
        \hline
  \multicolumn{8}{l}{Diffusion Methods}\\
        \midrule
        CLIP retrieval & 0.35 & 0.89 & 0.22 & 0.37 & 0.62 & 0.03 & 0.00 \\
        Stable Diffusion v1.5 & 0.43 & \underline{0.97}& 0.38 & 0.35 & 0.76 & 0.04 & 0.06 \\
        Stable Diffusion v2.1 & 0.50 & \textbf{0.98}& 0.51 & 0.44& \textbf{0.85}& 0.07 & 0.17 \\
        Stable Diffusion XL & 0.55 & \textbf{0.98}& \underline{0.74}& 0.39 & \textbf{0.85}& 0.15& 0.23 \\
        IF-XL & \textbf{0.61}& \underline{0.97}& \underline{0.74}& \textbf{0.66}& \underline{0.81}& 0.13 & \textbf{0.35}\\
        \hline
 \multicolumn{8}{l}{LLMs/MLLMs Enhanced Methods}\\
 \hline
 Llama-Gen&  0.32& 0.71& 0.34& 0.21& 0.58& 0.07&0.04\\
 SEED-X& 0.49& \underline{0.97}& 0.58& 0.26& 0.80& \underline{0.19}&0.14\\
 Emu3-Gen& 0.54& 0.98& 0.71& 0.34& \underline{0.81}& 0.17&0.21\\
 LWM& 0.47& 0.93& 0.41& \underline{0.46}& 0.79& 0.09&0.15\\
        \bottomrule
 \textbf{Ours-MEPT}& \underline{0.58}& \textbf{0.98}& \textbf{0.76}& 0.35& 0.80& \textbf{0.27}&\underline{0.29}\\
        \bottomrule
    \end{tabular}
\end{table*}
\noindent
\textbf{\subsection{4.2Evaluation Metrics}}
GenEval~\cite{ghosh2023geneval} is an object-level alignment benchmark specifically designed for evaluating Text-to-Image models. It overcomes the limitations of traditional metrics (such as FID and CLIPScore) by leveraging pre-trained object detection models to parse generated images, providing a detailed assessment of whether objects meet the attributes (such as color), quantity, spatial relationships, and co-occurrence relationships specified in the text description.
\begin{figure}
    \centering
    \includegraphics[width=1\linewidth]{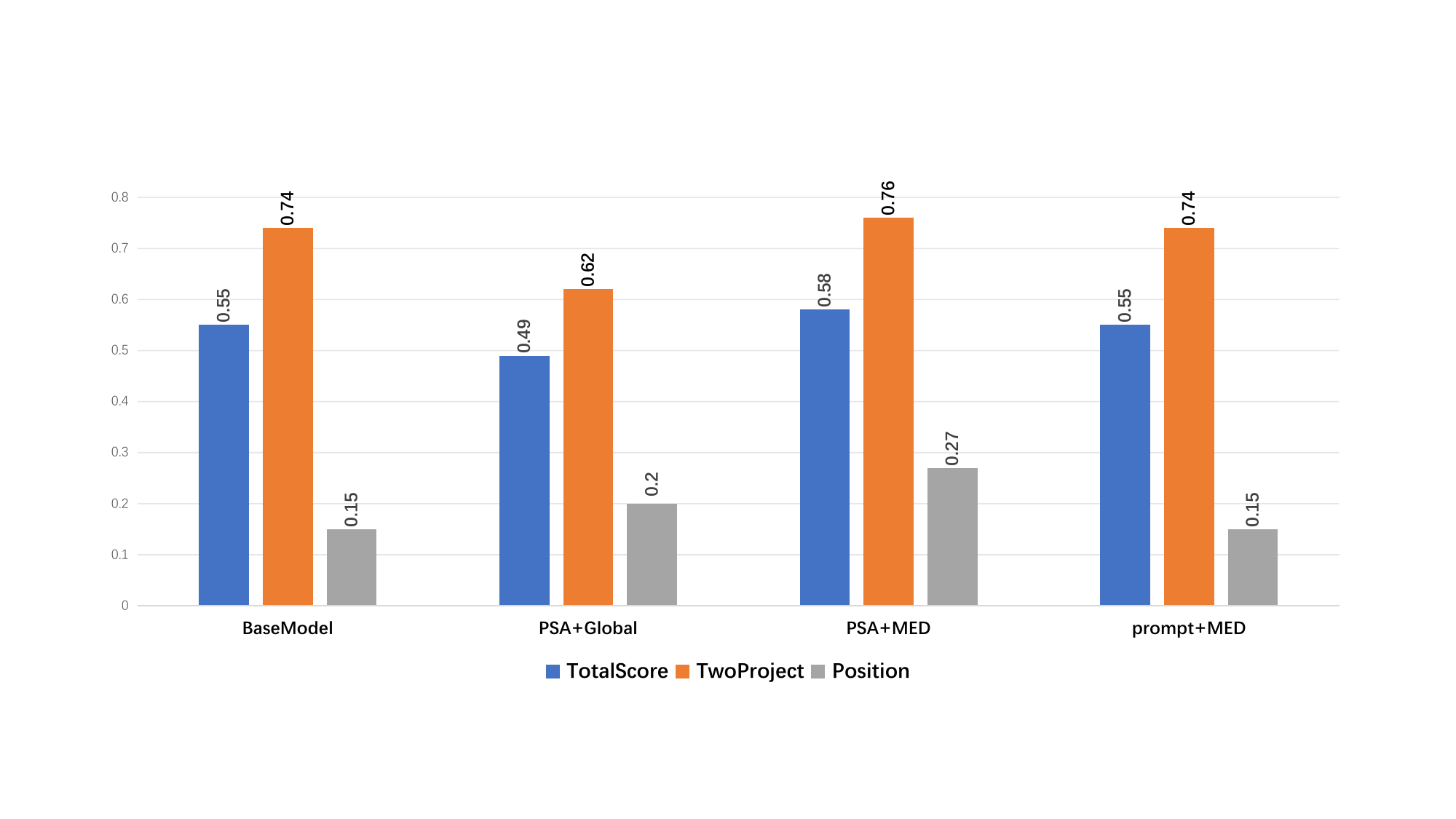}
    \caption{Ablation Mudules Experiment}
    \label{ablation}
\end{figure}
\begin{figure*}[htbp] 
\begin{center}
    \includegraphics[width=1\linewidth]{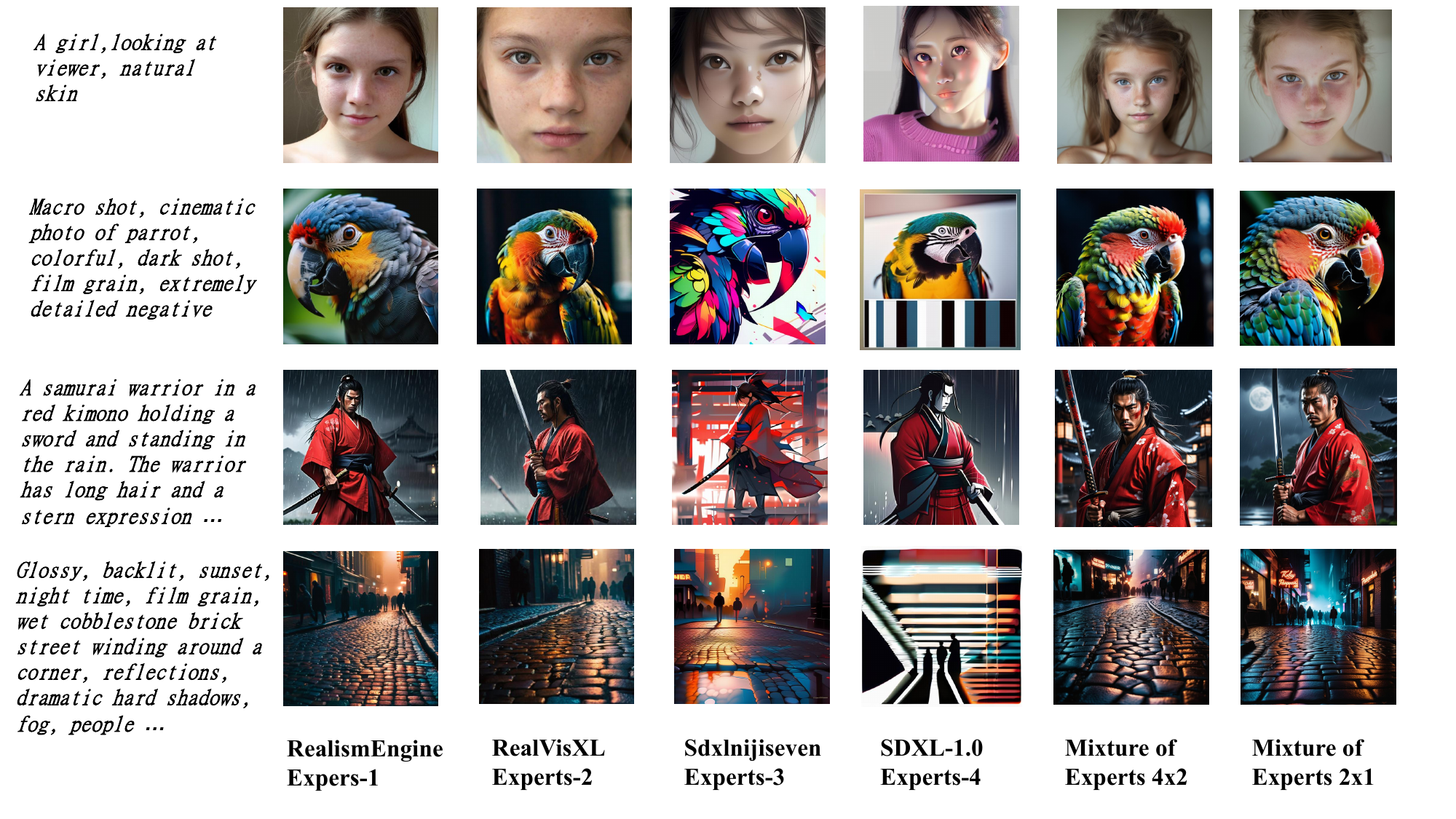}
\end{center}
    \caption{Comparison using different expert models and Mixture of Experts ( MOE) }
    \label{fig:6}
\end{figure*}
\vspace{0.2cm} %
\section{5 Results}\vspace{0.2cm} %

We conducted quantitative and qualitative evaluations of the Multi-Expert Planning and Generation Framework (MEPG)  framework in text-to-image generation. Experiments show that our reasoning-based generation method has improved in both complex multi-element text generation and style diversity. The ablation study also verified the effectiveness of our modules.   
\subsection{5.1 Quantitative Results }\vspace{0.2cm} %

Table 1 presents the evaluation of text-to-image generation (T-2-I) on Gen-Eval. The comparison covers two main types of models: one is the baseline-like Diffusion series models, and the other is models with LLM/MLLM enhance generation. In the T-2-I task, MEPT framework adopts $p1 = 0.5$. 
As shown in Table 1, our framework achieved the highest scores in both two-object generation (0.98) and position (0.27), and its total score is higher than that of the base model SDXL, proving the effectiveness of our MEPT framework. Although our model is slightly inferior to SDXL in counting tasks, our method outperforms in overall performance and more tasks, which indicates that the introduction of the MEPT framework can enhance the model's generation ability. Among the LLM/MLLM enhancement methods, our approach also outperforms in overall performance, demonstrating the rationality of the MEPT framework. 

\subsection{5.2 Qualitative Results}\vspace{0.2cm} %

In addition to the outstanding compositional text-to-image generation ability, MEPT also demonstrated the ability to diversify styles, as shown in Figure 6. We presented the effects of different expert-generated prompts. Our MOE model is proficient in generating various styles and can effectively generate coherent and aesthetically pleasing images for different prompts. 

\subsection{5.3 Ablation Studies}\vspace{0.2cm} %

We ablated the PSA modules and MED mudules of MEPT, and the results are shown in Figure 7. When only PSA is used, although the performance on the position task improves, the overall performance declines,This proves that PSA planning enhances the layout capability of the model, but the model requires a more effective denoising method to organically utilize this information. When PSA+MED is used, the overall performance has improved. Since our MED module requires information input with coordinates, when the default coordinates are (0,0), (1000,1000), the effect is the same as directly generated by the BaseModel. 
\subsection{5.4 Model Edit Generate}\vspace{0.2cm} %

In our experiments, we further demonstrated the manipulability of MEPT. As shown in Figure 7, this method enables users to modify the content of the PSA module to control the size, position, and style of image elements. 
 \begin{figure}[htbp]
    \centering
    \includegraphics[width=1\linewidth]{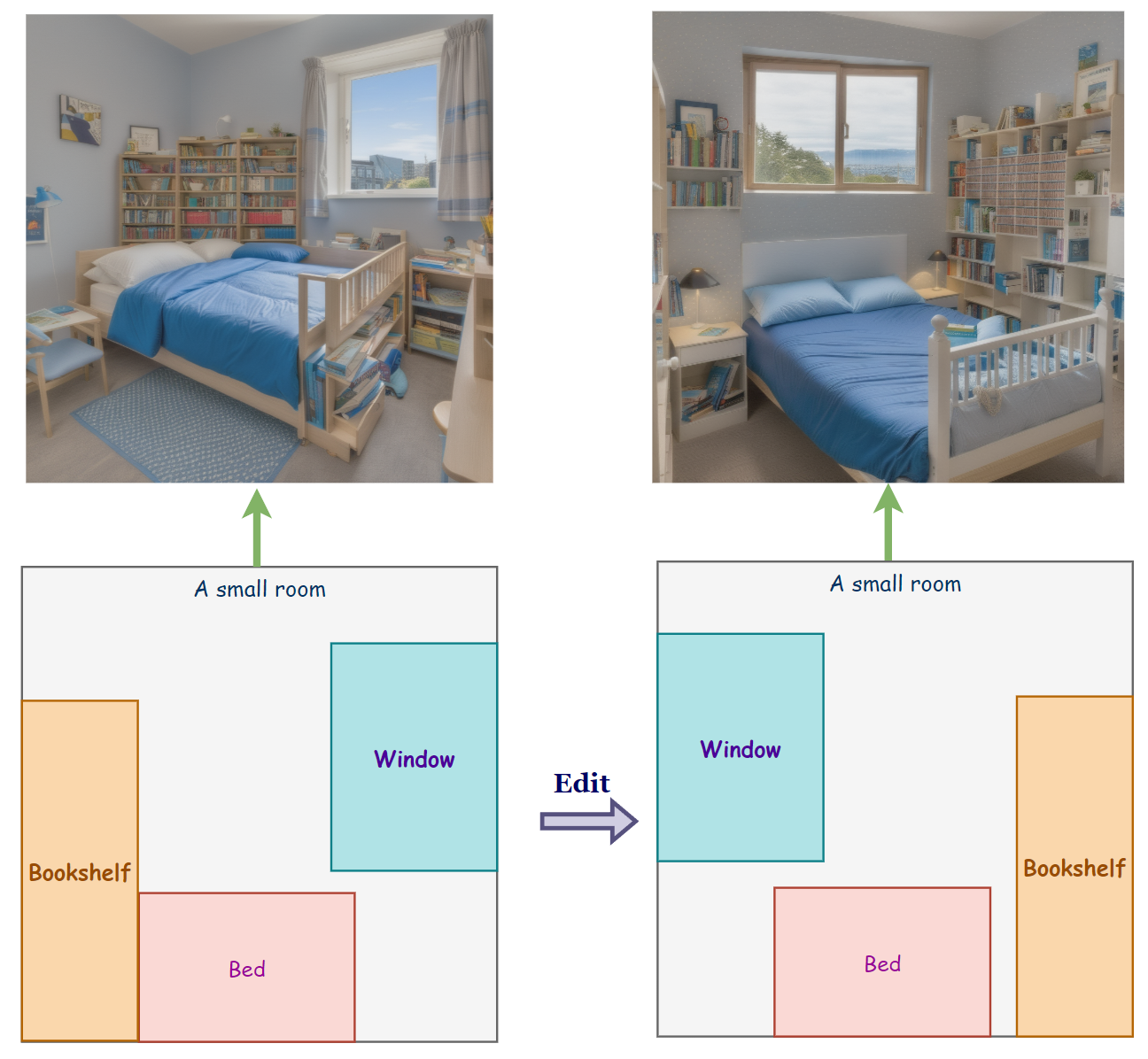}
    \caption{In this picture, we exchange the coordination between the window and bookshelf to get a different layout. }
    \label{fig:5}
\end{figure}

\section{6.Conclusion }\vspace{0.2cm} %
We proposed the Multi-Expert Planning and Generation (MEPG) Framework, which integrates the reasoning ability of large language models into image generation through the Position-Style-Aware (PSA) module and combines multi-expert models and spatial style generation through the Multi-Experts-Diffusion (MED) module. Our method transforms image generation from direct generation to a process with reasoning layout and style planning, and has precise spatial control capabilities, solving the problem that existing methods have difficulty generating complex, multi-element prompts and lack style diversity. At the same time, the lightweight expert module provides great potential for model scalability. Experiments prove that MEPT is a more systematic and powerful image generation framework. 

\section{7. Supplementary Material}\vspace{0.2cm} 

\subsection{Chain of Thought Design}

We have presented a part of our PSA code below, mainly about how to collaborate on information using LoRA and LLM:

\begin{lstlisting}[language=Python, frame=single, numbers=left, breaklines=true, basicstyle=\small]
def run_enhanced_chain(self, image_prompt):
    # Step 0: LoRA
    thought_str, lora_data = self.lora_initial_analysis(image_prompt)

    # Step 1
    elements = self.step1_find_main_elements(image_prompt, thought_str)
    position = self.step1_position_main_elements(image_prompt, elements)

    # Step 2
    coordinates = self.step2_arrange_elements(thought_str, elements, lora_data, position)

    # Step 3
    descriptions = self.step3_add_details(image_prompt, coordinates)

    try:
        transformed = self.transform_data_structure(descriptions, self.height, self.width)
    except Exception as e:
        transformed = self.transform_lora(lora_data, self.height, self.width)
        return transformed
    return transformed
\end{lstlisting}

We have also placed the more specific prompt design below, which includes LoRA prompt design and LLM prompt design:

\subsubsection{Lora}
\begin{lstlisting}[language=Python, frame=single, numbers=left, breaklines=true, basicstyle=\small]
"You are an expert in image understanding."
f"Image description: {raw_image} " 
f"Please list only the main objects or elements that may appear in the image. Elements are prohibited from being repeated."
f"Output only the names of the elements, separated by commas. Please pay attention to the exact quantity of the elements."
f"Limit the number of elements to at most {max_elements}."
"Do not output any extra content. Limited to 30 words."
"Main elements:"
\end{lstlisting}

\subsubsection{Steps}
\begin{figure}[ht]
    \centering
    \includegraphics[width=1\linewidth]{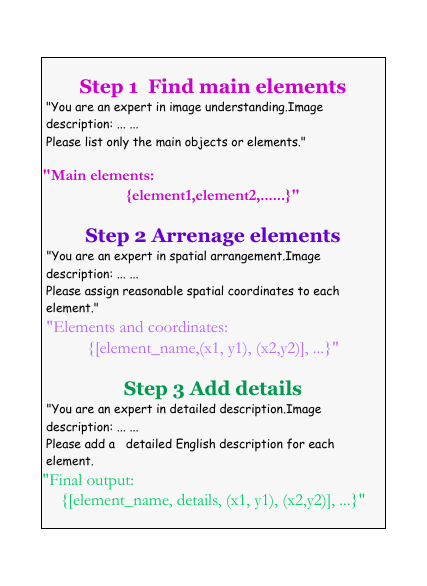}
    \caption{Experts Config}
    \label{fig:placeholder}
\end{figure}

\subsection{Multi-Experts Select}
We have selected four expert models as our basic models. Among them, Expert 1 and 2 are good at generating shooting styles, Expert 3 is good at generating anime styles, and Expert 4 is good at generating some characters.

\begin{lstlisting}[language=Python, frame=single, numbers=left, breaklines=true, basicstyle=\small]
experts:
  - expert1: 
      misri/realismEngineSDXL_v30VAE
  - expert2: 
      SG161222--RealVisXL_V5.0
  - expert3: 
      stablediffusionapi/sdxlnijiseven
  - expert4: 
      halcyon-sdxl-photorealism-v19-sdxl
\end{lstlisting}

\subsection{Other Code}
Our code will be made available on GitHub.


\bibliography{aaai25}

\end{document}